\documentclass[letterpaper, 10 pt, conference]{ieeeconf}  
\usepackage{graphicx}
\usepackage{subcaption}
\usepackage{amsmath}
\usepackage{todonotes}
\usepackage{algorithm}
\usepackage{algpseudocode}
\usepackage{tikz}
\usepackage{listofitems}
\usepackage{booktabs}
\usepackage{here}
\usepackage{url}
\usepackage{acronym}

\usepackage{etoolbox}

\renewcommand{\baselinestretch}{0.95}

\newtoggle{urs}
\togglefalse{urs}    

\newtoggle{icra}
\toggletrue{icra}

\newtoggle{doubleblind}
\togglefalse{doubleblind}

\tikzstyle{mynode}=[thick,draw=blue,fill=blue!20,circle,minimum size=22]
\tikzstyle{firstnode}=[thick,draw=green,fill=green!20,circle,minimum size=22]
\tikzstyle{lastnode}=[thick,draw=red,fill=red!20,circle,minimum size=22]

\IEEEoverridecommandlockouts                              

\newacro{FL}{federated learning}
\newacro{CADRE}{Cooperative Autonomous Distributed Robotic Exploration}
\newacro{NERFs}{Neural Radiance Fields}

\title{\LARGE \bf
Federated Multi-Agent Mapping for Planetary Exploration
}

\iftoggle{doubleblind}{\author{Anonymous Authors}}{
\author{Tiberiu-Ioan Szatmari$^{1,\dagger}$ and Abhishek Cauligi$^{2,\dagger}$
\thanks{$^{1}$Tiberiu-Ioan Szatmari, Department of Applied Mathematics and Computer Science, Technical University of Denmark \& Eriksholm Research Centre (email: {\tt\small tibs@dtu.dk}).} 
\thanks{$^{2}$Abhishek Cauligi is with the Department of Mechanical Engineering, Johns Hopkins University, Baltimore, MD USA (email: {\tt\small cauligi@jhu.edu}).}
\thanks{The research was carried out in part at the Jet Propulsion Laboratory, California
Institute of Technology, under a contract with the National Aeronautics and
Space Administration (80NM0018D0004).
}%
}
}

\begin{document}

\maketitle
\thispagestyle{empty}
\pagestyle{empty}

\begin{abstract}
\iftoggle{icra}{
Multi-agent robotic exploration stands to play an important role in space exploration as the next generation of robotic systems ventures to far-flung environments. A key challenge in this new paradigm will be to effectively share and utilize the vast amount of data generated onboard while operating in bandwidth-constrained regimes typical of space missions. Federated learning (FL) is a promising tool for bridging this gap. Drawing inspiration from the upcoming CADRE Lunar rover mission, we propose a federated multi-agent mapping approach that jointly trains a global map model across agents without transmitting raw data. Our method leverages implicit neural mapping to generate parsimonious, adaptable representations, reducing data transmission by up to 93.8\% compared to raw maps. Furthermore, we enhance this approach with meta-initialization on Earth-based traversability datasets to significantly accelerate map convergence—reducing iterations required to reach target performance by 80\% compared to random initialization. We demonstrate the efficacy of our approach on Martian terrains and glacier datasets, achieving downstream path planning F1 scores as high as 0.95 while outperforming on map reconstruction losses.
}{
Multi-agent robotic exploration stands to play an important role in space exploration as the next generation of spacecraft robotic systems venture to more extreme and far-flung environments. 
A key challenge in this new paradigm will be to effectively share and utilize the vast amount of data generated on-board while operating in bandwidth-constrained regimes such as those often found in space missions. 
Federated learning (FL) is a promising tool for bridging this gap for a host of tasks studied across proposed mission concepts.
Drawing inspiration from the upcoming CADRE Lunar rover mission, we propose a federated multi-agent mapping approach that jointly trains a global map model across agents without transmitting raw data. 
Our method leverages implicit neural mapping to generate parsimonious, adaptable representations, reducing data transmission by up to 93.8\% compared to raw maps. 
Furthermore, we enhance this approach with meta-initialization on Earth-based traversability datasets to significantly accelerate map convergence—reducing the number of iterations required to reach target performance by over 80\% compared to random initialization. 
We demonstrate the efficacy of our approach on Martian terrains and glacier datasets, achieving downstream path planning F1 scores as high as 0.95 while outperforming benchmarks on map reconstruction losses.
}
\end{abstract}

\section{INTRODUCTION}




In space exploration, the traditional paradigm of sending a single rover or spacecraft is giving way to new architectures that allow for faster, more efficient, and autonomous solutions that enable greater scientific returns and understanding.
For example, the Martian {\em Ingenuity} helicopter scouted terrain in advance for the {\em Perseverance} rover with the aim of identifying scientific targets of interest in advance and allowing for faster traverses for the rover~\cite{AndersonBrownEtAl2024}.
Indeed, such multi-agent mission concepts where teams of rovers or drones collaborate are poised to play a greater role in future exploration to more remote and far-flung planetary bodies in the Solar System~\cite{SchusterMuellerEtAl2020}.
A pressing challenge in enabling such mission architectures is that of information sharing, where the traditional concept-of-operations involves sending data generated on-board back to ground station operators on Earth to act upon.
This approach scales poorly with an increasing number of agents and can lead to costly time delays due to limited data transfer rates, both problems exacerbated by the increasing use of autonomy in planetary exploration~\cite{VermaMaimoneEtAl2023}.

\begin{figure}[t]
    \centering
    \includegraphics[width=.85\linewidth,height=.4\linewidth]{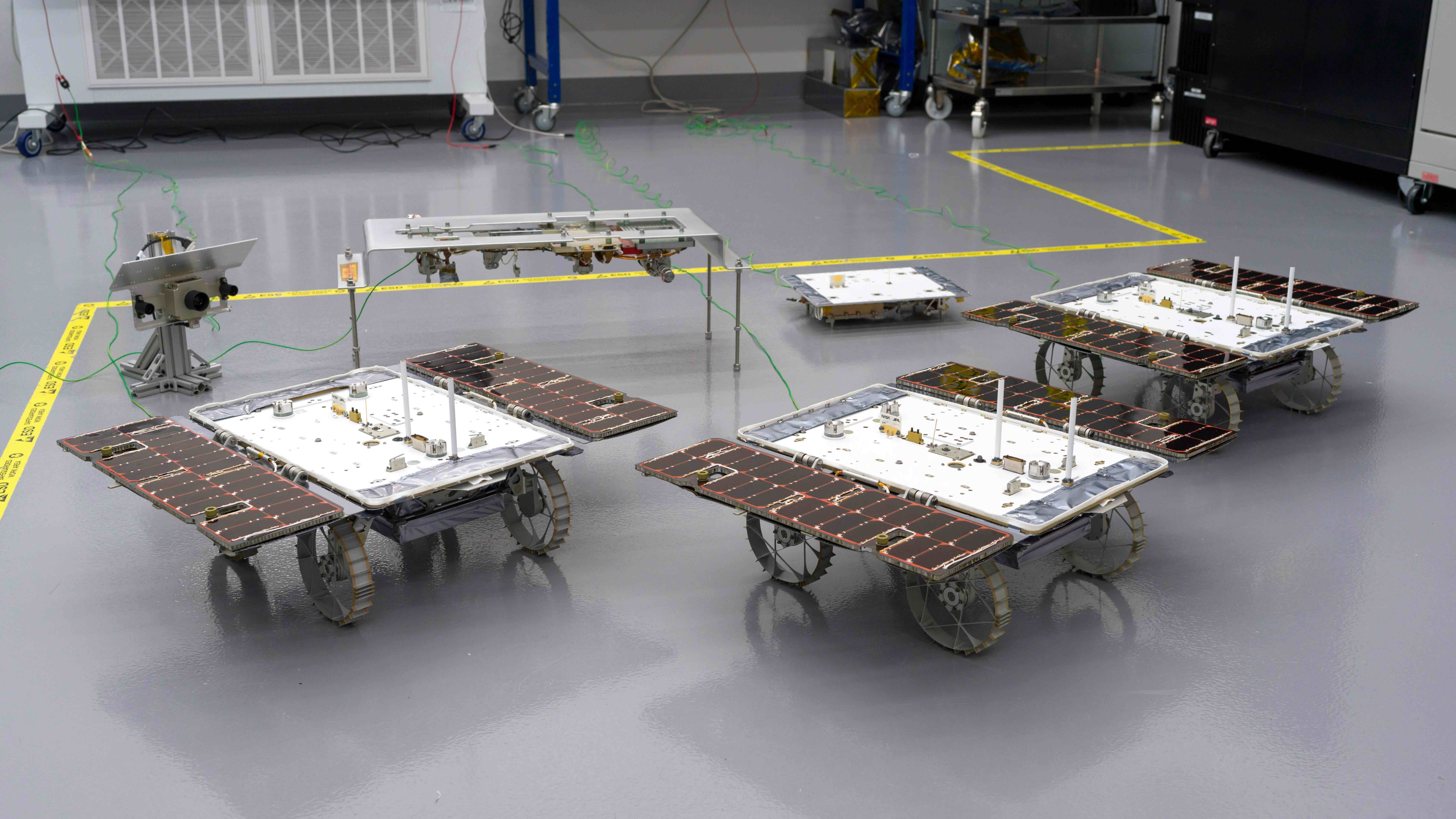}
    \caption{The CADRE mission, with the flight models shown here being tested in a clean room, is an upcoming multi-agent Lunar rover mission featuring three rovers performing distributed measurements of the Lunar surface and sub-surface, that aims to demonstrate how autonomy can dramatically transform scientific exploration capabilities across the Solar System~\cite{DeLaCroixRossiEtAl2024}.}
    \label{fig:cadre-rovers}
    \vspace{-2em}
\end{figure}

In this work, we draw inspiration from the field of~\ac{FL} and demonstrate how~\ac{FL} can enable teams of planetary robots to effectively carry out science and exploration tasks in the presence of such operating constraints.
\ac{FL} is a learning-based approach in which statistical models are trained in a distributed fashion without transmitting data directly~\cite{McMahanMooreEtAl2016, LiSahuEtAl2020}.
Although~\ac{FL} has become popular in medical and consumer applications due to privacy concerns that prevent the direct sharing of sensitive data, it has received limited attention for applications in space robotics and planetary exploration.
As such, we note that the upcoming~\ac{CADRE} mission ideally lends itself to the infusion of~\ac{FL} for the task of multi-agent distributed mapping. \ac{CADRE} is a Lunar rover mission in which a team of three Lunar rovers (Figure~\ref{fig:cadre-rovers}) map the Lunar environment by using distributed ground penetrating radar instruments~\cite{DeLaCroixRossiEtAl2024}.
During the Exploration phase of the mission, the three rovers will compute traversability maps on-board and transmit the raw map data to a central base station, where a global map of the environment is reconciled for use in science planning.

\begin{figure}[t]
    \centering
    \includegraphics[width=1\linewidth]{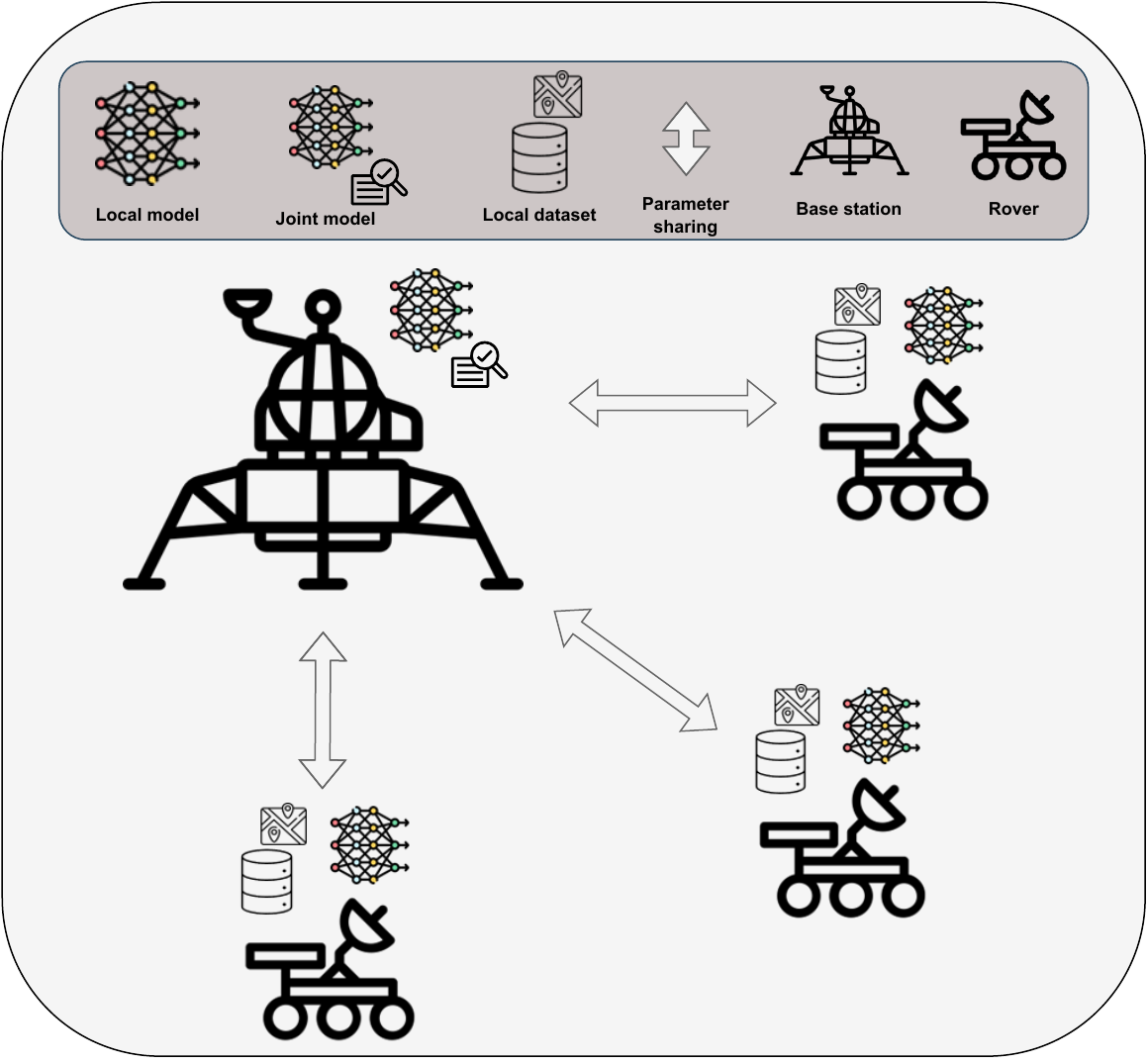}
    \caption{In future multi-agent space exploration, balancing communication efficiency with effective collaboration is paramount.  Traditional approaches that rely on sharing all raw data back to a base station quickly become infeasible due to bandwidth constraints. Federated learning addresses this by allowing rovers to learn maps and adapt their skills locally. They then share only the trained models with a base station, minimizing communication overhead. This enables efficient creation of a shared, global map representation while empowering individual rover autonomy.}
    \label{fig:cadre-art}
    \vspace{-2em}
\end{figure}

We propose a federated learning-based approach for tackling this problem of multi-agent mapping.
As shown in Figure~\ref{fig:cadre-art}, our approach targets multi-agent mission concepts such as~\ac{CADRE} where robots independently explore their surroundings and use on-board sensors to construct local traversability maps.
Each robot supervises a neural network online using data from its own local traversability map and key to our approach is the use of implicit mapping to train a specialized 2D version of~\ac{NERFs}~\cite{MildenHallSrinivasanEtAl2021}.
Global map construction is accomplished using federated learning, where only the trained neural network parameters from each robot are shared to the base station rather than the raw map data.
This global model is distributed back to the robots for continuous refinement, enabling effective map generation with few-shot or even one-shot communication rounds~\cite{ParkHanEtAl2021,ZhouPuEtAl2020}.

\iftoggle{icra}{}{
In this work, we propose a learning-based approach for distributed implicit mapping in multi-agent systems. Robots independently explore their surroundings using onboard sensors and their existing real-time mapping systems to create local maps. We represent these maps using a specialized 2D version of Neural Radiance Fields (NeRFs), meta-initialized on similar mapping data to accelerate the learning process. Importantly, our approach supports heterogeneous multi-agent systems where robots may have different sensor configurations and capabilities. Federated learning is employed to merge the trained NeRF parameters from each robot, creating a unified global map representation at a central base station. This global model is distributed back to the robots for continuous refinement, enabling effective map generation with few-shot or even one-shot communication rounds  \cite{ParkHanEtAl2021,ZhouPuEtAl2020}, particularly when combined with local post-processing.
}

To evaluate our approach, we leverage the availability of Earth-based mapping datasets for meta-training.
\iftoggle{icra}{}{
These datasets often originate from sources such as autonomous driving platforms~\cite{GeigerLenzEtAl2013,GeyerKassahunEtAl2020,SunKretzschmarEtAl2020}, aerial surveys~\cite{XiaHuEtAl2017}, and robotics research~\cite{TianChangEtAl2023}, providing a wide variety of terrain and environmental conditions.
}
This process establishes a robust prior for representing map-like data. We then rigorously test the ability of our method to adapt to out-of-distribution maps simulating diverse planetary surfaces.
Crucially, this demonstrates the efficacy of our proposed approach in generalizing to unforeseen and extreme environments.
For validation, we employ datasets from the Athabasca Glacier in Canada~\cite{ThakkerPatonEtAl2024}, with its challenging icy terrain, and the DoMars16k dataset~\cite{WilhelmGeisEtAl2020}, which includes simulated Martian geomorphological features from locations such as Jezero Crater (the Mars 2020 landing site~\cite{VermaMaimoneEtAl2023}) and Oxia Planum (the prospective ExoMars landing site).

\section{Related Work}\label{ch:related-work}
\subsection{Autonomy in Planetary Exploration}
The upcoming~\ac{CADRE} mission exemplifies the state-of-the-art in multi-agent, spacecraft robotic autonomy.
With the advent of low-power, high-performance computers such as the Qualcomm Snapdragon with integrated GPU in space robotics, missions such as~\ac{CADRE}~\cite{DeLaCroixRossiEtAl2024} are capable of generating high-resolution, local traversability maps fast enough for real-time operating rates~\cite{WernerProencaEtAl2024}. 

However, in a cooperative multi-agent mission, the situational awareness provided by each individual agent’s map is limited for science and mission planning.
Instead, map information must be shared between the agents to construct a global map, whereby valuable scientific measurement targets can be identified. 
For~\ac{CADRE}, ground station operators specify a region to be explored, after which exploration is performed autonomously.
Next, frontier candidates are computed from the shared traversability map and rovers are dynamically assigned to targets that minimize cost and maintain communication with the base.
This global map construction is accomplished using MoonDB~\cite{SaboiaRossiEtAl2024}, a centralized database on the mission base station that stores and fuses the traversability maps generated by each agent.
The key pitfall of MoonDB is that as the number of agents grow and each agent explores, the volume of data generated grows intractably and this bottlenecks global map construction. 

We argue that federated learning holds significant potential for addressing such challenges.
Indeed, recent years have demonstrated the efficacy of using~\ac{FL} to manage distributed data collected across teams of robots~\cite{XianjiaQueraltaEtAl2021}.
The authors in~\cite{NakanoyaImEtAl2021} apply~\ac{FL} for the task of training driver prediction models.
In~\cite{WangZhangEtAl2024}, the authors use~\ac{FL} to collaboratively train an imitation learning policy for robot grasping across a fleet of robots.
However, applying~\ac{FL} to space-borne applications presents unique challenges. The extreme and diverse range of environments encountered in space missions, coupled with limited prior knowledge about their structures, makes learning particularly difficult.

\subsection{Neural and Distributed Mapping} \label{sec:neur-and-distrib-map}
\ac{NERFs}~\cite{MildenHallSrinivasanEtAl2021} offer a way to create 3D representations from 2D images using deep neural networks to model volumetric density and color.
A NeRF function $F_{\theta} : (x, d) \rightarrow (c, \sigma)$ takes a 3D location $(x, y, z)$ and viewing direction $d$ as input, outputting radiance (color $c$) and density $\sigma$.
While NeRFs are used for 3D scene representation, a 2D version known as coordinate-based MLPs~\cite{SitzmannMartetEtAl2020} applies similar principles to map image coordinates $(x, y)$ to color value $c$ (RGB).

In neural mapping, networks are biased toward learning low-frequency features, limiting their ability to capture high-frequency details such as sharp textures~\cite{RahamanBaratinEtAl2019}. To address this, mapping inputs to higher-dimensional spaces improves the modeling of high-frequency variations \cite{TancikSrinivasanEtAl2020, MildenHallSrinivasanEtAl2021}.
Approaches such as RNR-Map~\cite{KwonParkEtAl2023} use NeRFs for 3D environment embedding, aiding localization and navigation, but struggle with unseen or ambiguous areas.
The federated learning approaches presented in~\cite{HoldenDayoubEtAl2023,Suzuki2023} reduce communication and training costs by collaboratively training NeRFs across distributed devices.
However, challenges remain in handling non-IID data, integrating diverse sensors, and adapting to dynamic environments, all of which are critical for multi-agent planetary exploration.

\section{Approach} \label{ch:approach}

\begin{figure*}[h]
    \centering
    \includegraphics[width=1\linewidth]{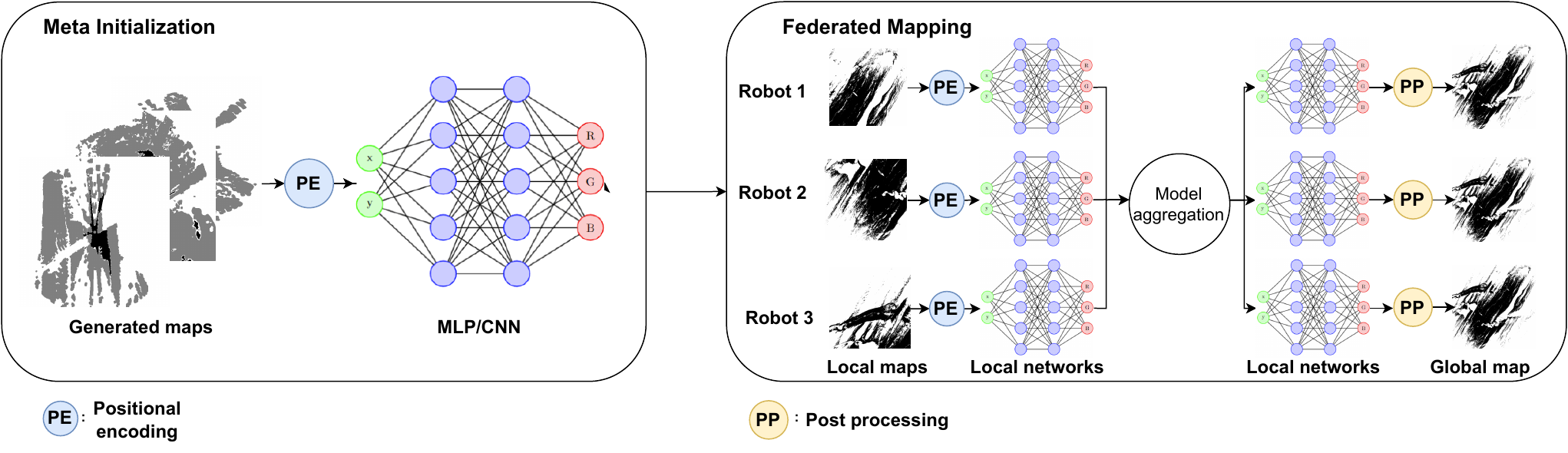}
    \caption{Our proposed federated mapping approach begins with an offline preparation stage where a neural network is trained on an empty (unknown) grid map and then meta-trained with Reptile on a map dataset for quick adaptation (this is done only once, offline). Next, multiple agents explore within a global reference frame, generating local area maps. Each agent utilizes a neural network to learn its local map. These learned network parameters are then shared with a central server where model aggregation occurs. Finally, the updated joint model parameters, now containing global map knowledge, are distributed back to each robot. Locally, robots can further refine (e.g. remove noise) the global map generated from the updated model.}
    \label{fig:detail-diagram}
    \vspace{-2em}
\end{figure*}

\subsection{Overview}
To address the challenges described in Section ~\ref{ch:related-work}, we introduce a federated learning framework for multi-agent planetary exploration. Each agent learns local maps using 2D networks, with offline meta-training on Earth-based data. This establishes a robust prior that enables networks to rapidly learn from limited local data and viewpoints, addressing the challenges of non-IID data and ensuring local model performance, while supporting heterogeneous sensor configurations. Summarizing our contributions:

\begin{enumerate} 
\item \textbf{Offline Meta-training}: Accelerates network adaptability to new maps via meta-training on traversability datasets. 
\item \textbf{Collaborative Map Building}: Agents build a global map by exchanging learned network parameters, minimizing communication overhead. 
\item \textbf{Map Evaluation \& Refinement}: Validates refined mapping performance via image quality metrics (PSNR, SSIM) and path planning simulations (F1).
\end{enumerate}

Our map refinement process uses morphological image processing (e.g. connected elements) to fill gaps and remove noise artifacts, improving map clarity. Fig.~\ref{fig:detail-diagram} illustrates the approach and Alg.~\ref{alg:fed-map-meta} provides high-level pseudocode.

\subsection{Mapping the Unknown} \label{sec:map-unk}

In multi-agent space exploration, robots gather data within limited local views, resulting in incomplete global maps. Learning from this partial information can lead to inaccuracies, especially in unexplored areas. Moreover, raw coordinates $(x,y)$ are insufficient for networks to capture the fine spatial variations needed for accurate map representation in planetary environments.

To address these challenges, we use a Gaussian encoding inspired by Fourier Feature Networks \cite{TancikSrinivasanEtAl2020}, transforming $(x,y)$ coordinates into a higher-dimensional representation (Eq. \ref{eq:pos-enc}), allowing the network to capture the intricate spatial details essential for space exploration. We also pre-train the network on an image representing an unexplored map (e.g., a blank image) to prevent the misinterpretation of unseen regions, ensuring the map updates accurately as data becomes available. This step, unlike meta-initialization focused on rapid adaptation, focuses on mitigating artifacts caused by the reference frames common in space missions.

The network takes positionally encoded coordinates as input and produces a map as output, where pixel values represent learned traversability information. This encoding is given by:
\begin{align}
    \gamma(v) = 
    \begin{bmatrix} 
    \cos(2\pi Bv) \\ 
    \sin(2\pi Bv) \\ 
    \end{bmatrix},
    \label{eq:pos-enc}
\end{align}

where $v$ represents the input coordinates $(x,y)$ and $B$ is a matrix of random frequencies from a Gaussian distribution. The scale parameter, set to 10, adjusts the frequency matrix $B$, enhancing the network’s ability to capture finer spatial details. We use two input channels and a mapping size of 128 to define the feature space for space exploration tasks.

\subsection{Meta-initialization}

To accelerate map learning and improve generalization for diverse terrains in space, we use a meta-initialization strategy. First, we train a neural network on the previously described empty map with encoded grid inputs. Then, using the Reptile meta-learning framework~\cite{NicholSchulman2018}, we train this network on Earth-based traversability and elevation maps, such as those from the KITTI dataset~\cite{GeigerLenzEtAl2013}. We treat different categories in the dataset (urban, rural, highways, etc.) as separate tasks for meta-learning, enabling the model to generalize to new terrains. This meta-learning process aims to establish a strong prior for representing diverse map-like data, anticipating the variety of terrains encountered in space missions. Inspired by~\cite{TancikMildenHallEtAl2021}, this learned initialization leads to faster adaptation to new, out-of-distribution maps.

Reptile, a first-order meta-learning algorithm, finds model initialization parameters that allow rapid adaptation with few updates. It works by sampling tasks, performing gradient steps, and updating the initialization based on the difference between initial and task-updated parameters. The update rule can be expressed as:
\vspace{-0.5em}
\begin{align}
    \theta \leftarrow \theta - \epsilon \sum_T ( \theta - \nabla_{\theta} L_T ( \phi_T)),
\end{align}
where $\theta$ represents the model parameters, $T$ is a sampled task, and $L_T$ is the task loss. $\epsilon$ is the step size, and $\phi_T$ represents task-specific updated parameters.

\subsection{Network Definition}~\label{sub:net-def}
 The network is defined as a simple convolutional architecture for 2D implicit mapping, with the default configuration consisting of four convolutional layers (kernel size 1, no padding) and an output of 3 (RGB), using ReLU activations \cite{Agarap2018} and batch normalization. This configuration functions similar to a fully-connected (dense) layer, making the network resemble a multi-layer perceptron (MLP). The architecture is scalable, allowing flexibility from simple linear layers to more complex convolutional ones. The parameter count is fixed, depending only on the input and output channels, not input resolution, making it independent of spatial dimensions.
The network uses 256 channels to match positional encoding and is trained using the ADAM optimizer~\cite{KingmaBa2015} with Mean Squared Error (MSE) loss.

\subsection{Federated Mapping}
Using the network from \ref{sub:net-def}, our federated mapping process starts with learning an unexplored grid (Line \ref{alg:empty-map}) and meta-training (Line \ref{alg:meta-train}), to enable swift adaptation. Each agent is initialized with meta-trained parameters (Line \ref{alg:meta-init-agent}).
Agents explore independently, generating local maps (Line~\ref{alg:explore-and-map}) and training the network using the ADAM optimizer and MSE loss (Line \ref{alg:learn-map}). Trained parameters are sent to the central server (Line~\ref{alg:centralize}) for aggregation (e.g., FedAvg) to create an updated global model (Line~\ref{alg:fed-agg}), which is redistributed to agents (Line~\ref{alg:fed-update}). The global model is then used to generate a global map at each agent (Line~\ref{alg:generate-global-map}), followed by gap filling and artifact removal (Line~\ref{alg:process-global-map}).
 Importantly, we focus on \textit{few-shot} federated learning, aiming for effective map refinement with minimal communication rounds, iterating as agents explore.

\subsection{Federated Map Refinement}
In our one-shot scenario, the federated approach captures the overall environmental structure in a single communication round, minimizing data transmission between agents and the central server—a key advantage for bandwidth-limited space missions. To improve the global map, we apply efficient map refinement techniques (Fig. \ref{fig:fed-map}, Subfig. 6).
We first scan the map to identify small gaps in navigable areas caused by sensor noise or imperfections and fill them by analyzing neighboring pixels (majority vote). This ensures continuous navigable regions for smooth path planning.
Next, we remove small isolated obstacles or noise, identifying and eliminating minor pixel clusters below a certain threshold (e.g. 200) to prevent misinterpretation by path planning algorithms. 
These thresholds, determined through experiments on map data and fixed across our tests, reliably produce hole-free maps for downstream path planning algorithms. Should planetary terrain variations require different settings, these parameters can be dynamically adjusted.


\begin{algorithm}[t]
\caption{Federated Mapping with Meta-initialization}
\label{alg:fed-map-meta}
\begin{algorithmic}[1]
\Require{$M_0$ an empty map, $M_\textrm{global}$ the federated global map, $D_\textrm{trav}$ traversability map dataset, $F$ global reference frame}

\Statex \textbf{Offline Meta-Training}
\State Train a network on the empty map
\State $\theta_0 \leftarrow \textrm{Train}(\textrm{Net}, M_0)$  \label{alg:empty-map}
\State $\theta \leftarrow \textrm{MetaTrain}(\textrm{Net}, \theta_0, D_\textrm{trav})$  \label{alg:meta-train}
\newline
\Statex \textbf{Federated Map Learning}
\State \textbf{Initialize:} Distribute meta-trained network parameters $\theta$  \label{alg:meta-init-agent}
\For{each agent i in parallel}
\State Explore local area, generating map $M_i$  \label{alg:explore-and-map}
\State $\theta_i \leftarrow \arg\min_{\theta} L(M_i, \textrm{Net}(\theta))$  \Comment{Local network training}  \label{alg:learn-map}
\State Send $\theta_i$ to the central server  \label{alg:centralize}
\EndFor
\State $\theta_\textrm{global} \gets \frac{1}{N} \sum_{i=1}^N \frac{n_k}{n}\theta_i $ \Comment{Federated Aggregation}  \label{alg:fed-agg}
\State Distribute $\theta_\mathrm{global}$ to each agent  \label{alg:fed-update}
\State $M_\textrm{global} \leftarrow \textrm{Net}(\theta_\textrm{global})$ \Comment{Generate global map} \label{alg:generate-global-map}
\State $M_\textrm{global} \leftarrow \textrm{RefineMap}(M_\textrm{global})$ \Comment{Filter and remove noise} \label{alg:process-global-map}

\State \textbf{Repeat} exploration, training, and aggregation as needed

\end{algorithmic}
\end{algorithm}

\section{Experiments} \label{ch:experiments}

\begin{figure*}[t]
    \includegraphics[width=1\linewidth]{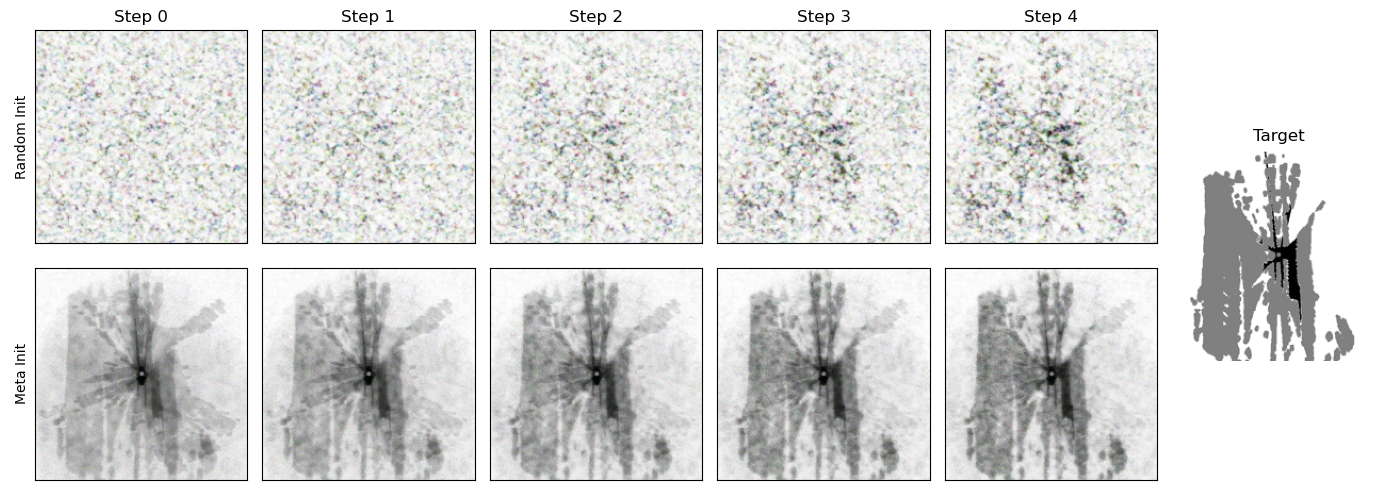}
    \caption{Meta-initialization with Reptile enables faster network convergence for map reconstruction from KITTI data (far right). This is demonstrated by comparing a randomly initialized model (top) and its meta-initialized counterpart (bottom) from steps 0-4. Each step represents two optimization iterations.}
    \label{fig:meta-init-kitti}
    \vspace{-1em}
\end{figure*}
\subsection{Datasets}
\begin{figure}[t]
    \centering
    \begin{subfigure}{0.2\textwidth}
        \centering
        \includegraphics[width=\linewidth]{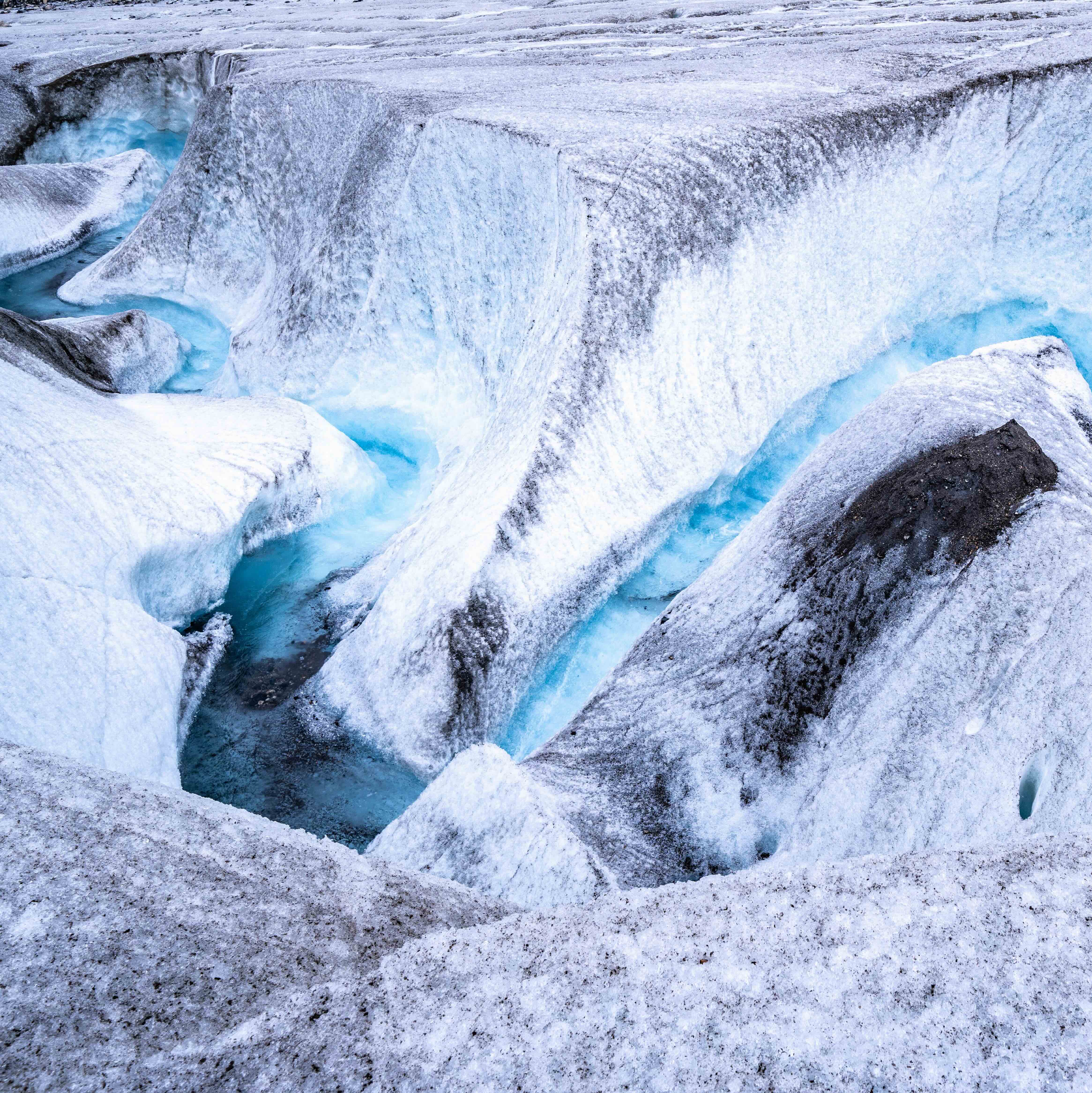}
        \caption{Athabasca Glacier landscape.}
        \label{fig:ath-3d}
    \end{subfigure}
    \hfill
    \begin{subfigure}{0.2\textwidth}
        \centering
        \includegraphics[width=\linewidth]{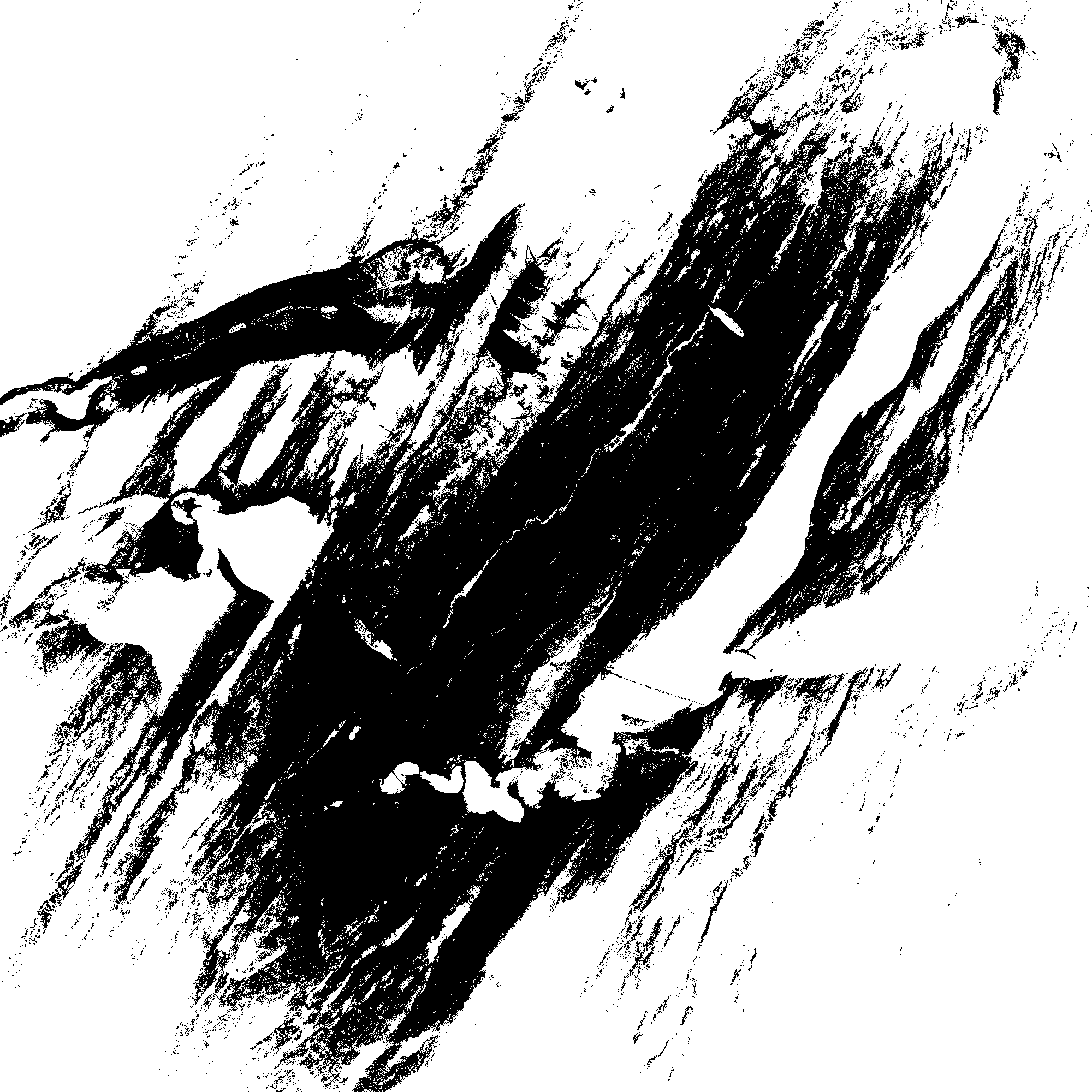}
        \caption{Traversability map (black: clear)}
        \label{fig:ath-map}
    \end{subfigure}
    \caption{Athabasca Glacier: from landscape picture to map representation.}
    \label{fig:athabasca-example}
    \vspace{-2em}
\end{figure}
Our evaluation strategy combines real-world Earth-based datasets for meta-initialization with simulated planetary environments for testing out-of-distribution adaptation.

For meta-initialization, we leverage the KITTI dataset~\cite{GeigerLenzEtAl2013}, which includes traversability and elevation maps from urban and rural settings around Karlsruhe, Germany. KITTI provides diverse data collected using high-resolution cameras, laser scanners, and GPS. We treat different categories (Residential, Road) as separate tasks for meta-learning, creating a prior for varied terrains. While the specific categories do not occur on the Moon, they capture a broad range of surface variations—such as different textures, elevation changes, and traversability cues—that provide a robust, diverse prior.  We use a train/test/validation split of 2,385, 282, and 141 map samples.

For adaptability testing, we utilize the unique dataset collected by the Exobiology Extant Life Surveyor (EELS)~\cite{ThakkerPatonEtAl2024} project at JPL. EELS focuses on developing ``snake-like'' autonomous robots to explore icy planetary bodies, a nascent area in space exploration. The dataset, collected at the Athabasca glacier in Canada, provides a large-scale point cloud converted into 2D traversability and elevation maps. The vast ice field of the glacier and crevasses simulate the extreme conditions anticipated on Europa, an icy moon of Jupiter, offering a distinctive proving ground for evaluating our approach’s ability to handle the challenges of icy world exploration.
Additionally, the DoMars16k dataset~\cite{WilhelmGeisEtAl2020} simulates a diverse range of Martian features, such as craters and dunes, providing a testbed for assessing our system’s generalization to the unique terrain of Mars.

This dataset strategy offers two major advantages.  First, we leverage the abundance of Earth-based data for robust initialization. Second, we rigorously test the federated system's ability to quickly adapt to new and potentially different environments to be encountered in future space exploration.

\begin{table}[t]
\centering

\caption{Comparative analysis of initialization methods for map learning. We report the PSNR after two iterations of test optimization, showing that the meta-learned initialization (Meta) outperforms other methods. Additionally, we present the avg. number of iterations required for the other methods to reach the Meta's performance level after two iterations.}
\label{tab:meta-init}
\begin{tabular}{@{}lcc@{}}
\toprule
\textbf{Initialization} & \textbf{PSNR $\uparrow$} & \# \textbf{iter to match $\downarrow$} \\ \midrule
Random&  8.16& 68.68 $\pm$ 1.06\\
Empty map& 9.37& 26.98 $\pm$ 1.84 \\
\textbf{Meta} & \textbf{13.30} & \textbf{\textemdash} \\ \bottomrule
\end{tabular}
\end{table}
\subsection{Metrics}

We employ two main image quality metrics to evaluate our federated mapping approach. PSNR measures pixel-level differences between the reconstructed and ground truth maps, with higher values indicating better reconstruction quality. However, PSNR may not fully capture map accuracy for navigation tasks. SSIM addresses this by considering contrast, and structure, with values ranging from -1 to 1, where 1 indicates perfect similarity.


We evaluate path planning by comparing A* paths on true versus learned maps. A true positive is counted when the paths are identical.
A false positive occurs when the learned map yields a path where none exists in the ground truth and a false negative when it fails to find one that does.
This strict setup tests whether the map supports the same optimal path, offering a strong measure of navigational fidelity, similar to prior work~\cite{qian2022hy}.

\subsection{Meta-initialization Ablation Results}

In Fig.~\ref{fig:meta-init-kitti}, we compare meta-initialization with random initialization after 1000 iterations and a learning rate of $10^{-4}$ across $200$ test samples per run. The meta-initialized network quickly captures fine-grained map details, while the randomly initialized one takes significantly longer to reach similar fidelity. As shown in Table~\ref{tab:meta-init}, the meta-learned network, initiated from an empty map, achieves an average PSNR of 13.30 after just one learning step (two forward passes) on novel test maps, outperforming random initialization (8.16 PSNR) and empty map initialization (9.37 PSNR). Random initialization requires 68.68 ± 1.06 forward passes, and empty map initialization needs 26.98 ± 1.84 to match the meta-trained network’s PSNR, demonstrating faster adaptation.
Thus, we see that meta-initialization using readily available Earth-based datasets enables performant~\ac{FL} and fast adaptation for space-based applications.

\subsection{Federated Mapping Results}
We evaluated three federated learning algorithms—FedAvg~\cite{McMahanMooreEtAl2017}, FedAdam~\cite{ReddiCharlesEtAl2021} and FedYogi~\cite{ReddiCharlesEtAl2021}-on multi-agent mapping in planetary exploration, focusing on their ability to handle heterogeneous, non-IID data typical of these environments. 
We hypothesized that FedAvg would outperform adaptive optimizers such as FedAdam and FedYogi due to challenges posed by the non-uniform data distribution across agents. 
FedAdam~\cite{KingmaBa2015} and FedYogi~\cite{zaheer2018adaptive} were selected to explore the potential benefits of adaptive learning rates, which adjust based on the first and second moments of the gradients. 
While these optimizers could theoretically offer faster convergence, they may struggle in highly diverse environments due to instability during aggregation.

\begin{figure*}[t]
  \centering
  \begin{subfigure}{\linewidth}
    \includegraphics[trim=0cm 1cm 0cm 0cm, clip,width=1\linewidth]{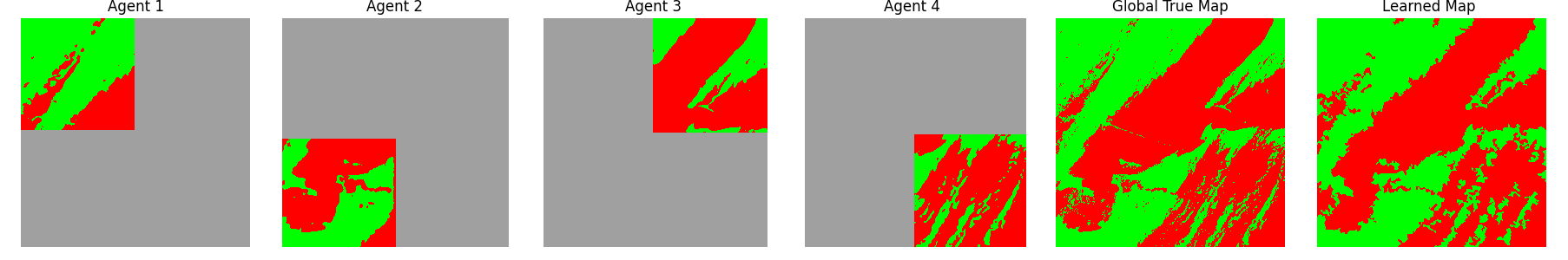}
    \caption{Federated mapping on the Athabasca Glacier. Progression described in the figure caption.}
    \label{fig:fed-map-a} 
  \end{subfigure}
  \vspace{0.1cm} 
  \begin{subfigure}{\linewidth}
    \includegraphics[width=1\linewidth]{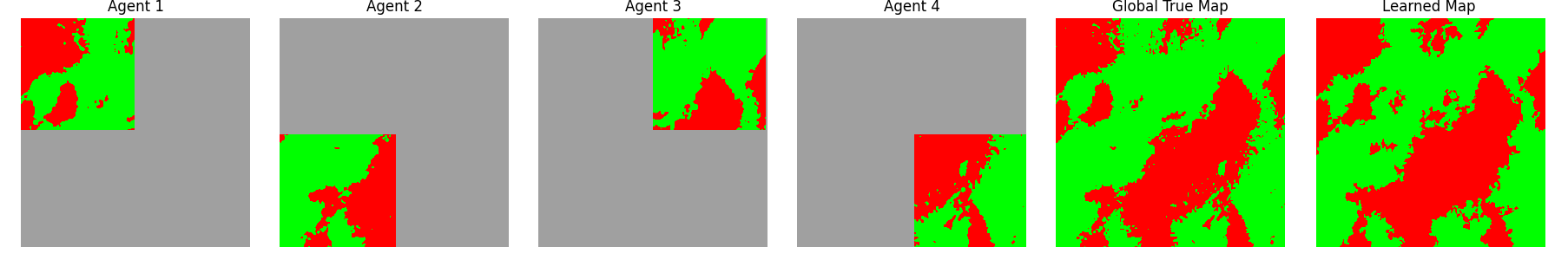}
    \caption{Federated mapping on a Martian surface.} 
    \label{fig:fed-map-b} 
  \end{subfigure}
  \caption{Federated map learning results for both the Athabasca Glacier (top row) and a Martian surface (bottom row). For both, the figure progresses as follows: the first four subfigures display the local maps learned by individual agents. Subfigure 5 shows the ground truth global map. Subfigure 6 presents the final global map obtained by combining the agent maps through federated learning. \textcolor{green}{Green} = free; \textcolor{red}{Red} = impassable; \textcolor{gray}{Gray} = unknown.} 
  \label{fig:fed-map}
  \vspace{-1em}
\end{figure*}

To test this hypothesis, we used two out-of-distribution datasets: the Athabasca Glacier and a simulated Martian surface.
As illustrated in Fig.~\ref{fig:fed-map}, four agents explored distinct areas, learning the generated local maps and sharing trained parameters in a single communication round ($R=1$). Table~\ref{tab:fed-map} provides a comprehensive evaluation of one-shot mapping performance across key metrics: PSNR (map reconstruction quality), SSIM (structural similarity) and F1 scores (path planning efficacy), using consistent settings (4 agents, 100 local epochs). 

\begin{table}[t]
\centering
\caption{Federated mapping results on out-of-distribution datasets (Athabasca Glacier and Mars Surface) against ground truth, with performance assessed through PSNR, SSIM, and the F1 score, which is directly tied to path planning efficacy across different algorithms.}
\label{tab:fed-map}
\begin{tabular}{@{}lcccccc@{}}
\toprule
 & \multicolumn{3}{c}{\textbf{Athabasca Glacier}} & \multicolumn{3}{c}{\textbf{Mars Surface}} \\ \midrule
\textbf{Algorithm} & PSNR $\uparrow$  & SSIM $\uparrow$  & F1 $\uparrow$ & PSNR $\uparrow$  & SSIM $\uparrow$  & F1 $\uparrow$ \\ \midrule
FedAdam & 5.27 & 0.31 & 0.1 & 4.4 & 0.25 & 0.03 \\
FedYogi & 10.73 & 0.44 & 0.42 & 9.19 & 0.39 & 0.87 \\
FedAvg & \textbf{10.85} & \textbf{0.47} & \textbf{0.95} & \textbf{9.26} & \textbf{0.42} & \textbf{0.94} \\
\bottomrule
\end{tabular}
\end{table}
(1) FedAdam integrates Adam's adaptive learning rates into federated learning but performs poorly in our experiments, with low PSNR, SSIM, and near-zero F1 scores. 
The main issue is likely its sensitivity to gradient differences across agents. In highly heterogeneous surfaces such as the glacier crevasses, where local data distributions differ significantly, the adaptive adjustments seem to have caused instability during aggregation, leading to excessive or insufficient learning rate adjustments for effective global models. 
(2) FedYogi, an adaptation of the Yogi optimizer for federated settings, improves over FedAdam, especially on the Mars dataset with an F1 score of 0.87. However, it still struggled with the high data heterogeneity, as the learning rate adjustments were still too reactive to the large variations in local environments. 
(3) FedAvg outperforms both adaptive methods, achieving the highest PSNR (10.85), SSIM (0.47), and F1 scores (0.95 on Athabasca Glacier, 0.94 on Mars). These results suggest that FedAvg's simpler approach of averaging local updates, without dynamic learning rate adjustments, proved more stable in the tested environments, leading to superior map reconstruction and path planning across diverse terrains. In contrast, adaptive methods such as FedAdam and FedYogi struggled with instability during model aggregation, rendering them less effective for federated learning in multi-agent planetary exploration.

\subsection{Data Transfer Comparison}
Our federated mapping approach significantly reduces data transmission size compared to sending raw map data.
Table~\ref{tab:data-comparison} compares raw data sizes with the neural network parameters transmitted in our method.
The network has approximately 200 thousand parameters: 65,792 per layer for Layers 1-3 (including biases), 512 per BatchNorm layer, and 771 for the final layer. Each parameter is stored as a 16-bit (2-byte) value, making the total model size approximately 399 KB, with the potential for further quantization.

\begin{table}[t]
\centering
\caption{Comparison of Data Transmission Sizes between Raw Map Data and Model Parameters}
\label{tab:data-comparison}
\begin{tabular}{@{}lccc@{}}
\toprule
\textbf{Map Resolution} & \textbf{Data Size} & \textbf{Model Size} & \textbf{Reduction} \\
\midrule
\multicolumn{4}{l}{\textit{Grayscale Map}} \\
2,000 $\times$ 2,000 pixels & 3.81 MB & 399 KB  & 89.53 \% \\
- As JPEG (Q=26) & 402 KB & 399 KB  & 0.7\% \\
\midrule
\multicolumn{4}{l}{\textit{OMG Map}} \\
400 $\times$ 400 cells    & 960 KB   & 399 KB  & 58.4 \% 
\\
600 $\times$ 600 cells    & 2.16 MB   & 399 KB  & 81.5 \% \\
\midrule
\multicolumn{4}{l}{\textit{CADRE CCM}} \\
400 $\times$ 400 cells    & 2.88 MB   & 399 KB  & 86.1 \%
\\
600 $\times$ 600 cells    & 6.48 MB   & 399 KB  & 93.8 \% 
\\
\bottomrule
\end{tabular}
\vspace{-2em}
\end{table}

Transmitting the full model parameters requires approx. 399 KB. This is already more bandwidth-efficient than sending raw map data for any instance of the grayscale, optimal mixture of gaussians (OMG) map proposed for Mars rotorcrafts~\cite{proencca2022optimizing}, or the converging covariance map (CCM) used for CADRE~\cite{DeLaCroixRossiEtAl2024}.
In particular, we see that the CADRE CCM requires approximately 2.88 MB for a 400$\times$400 cell map (18 bytes/cell).
We note that this comparison is for a single CCM map, whereas the Exploration phase of the CADRE mission entails computing multiple CCMs as the rover drives and each one of these CCM instances is transmitted back to the base station for map merging.
In addition, JPEG with quality factor (Q=26) can perform lossy compression for a raw grayscale map (2000×2000) to 402 kB -nearly matching the full model parameters. 
In this configuration, it can yield high quality metrics (PSNR: 25.88, SSIM: 0.90), and a similar downstream path planning F1 score of 0.96. 

However, our approach stands apart from conventional image compression such as JPEG because the model can leverage gradient sparsification. Techniques such as top-k selection, as demonstrated in Deep Gradient Compression (DGC)~\cite{lin2017deep}, can reduce the data transmitted by roughly 99\% compared to full gradient updates for CNN-based ResNet architectures. 
In our task, where the full model is approximately 399 kB, this implies that only about 1\%—roughly 4 kB—needs to be sent per update round. 
This dramatic reduction in update size significantly outperforms traditional JPEG compression. 
Moreover, the meta-initialization strategy functions similarly to the warm-up training phase in DGC. 
This warm-up establishes a strong prior, such that early updates are less noisy and more stable. 

\subsection{Downstream Path Planning}
We evaluated the practical utility of our federated mapping by running A* path planning simulations on 75 randomly selected routes from real terrain data. 
In this context, a path refers to the route calculated by the A* algorithm, connecting a validated start point to an end point on either the true or learned map. 
Each path consists of a sequence of navigable waypoints that the robot follows while avoiding obstacles and traversing safe terrain. 
We compared the paths generated from the learned global maps against 
ground-truth traversability maps.
By comparing these paths, we assessed how accurately the learned maps supported feasible route planning. 
Precision, recall, and F1 score were used to quantify the model's ability to identify navigable paths and avoid obstacles.
While PSNR and SSIM indicated map quality, the F1 score was key in evaluating path planning effectiveness, showing a strong correlation with improved path planning and safer navigation.

These results demonstrate the benefits of federated multi-agent mapping, even in a one-shot scenario with minimal communication. The integration of local maps into a global traversability map enables reliable autonomous navigation, crucial for bandwidth-limited space missions.

\section{Conclusion} \label{ch:conclusion}
In this paper, we introduce a federated multi-agent mapping strategy for space exploration robotics, utilizing implicit neural mapping for efficient and detailed map creation.
This method supports rapid adaptation in diverse environments through meta-initialization and collaborative learning.
Validation on mission-like datasets confirms its effectiveness.
\iftoggle{icra}{
In future work, we aim to apply~\ac{FL} for mission concepts in which information must be shared across heterogeneous agents (e.g., a rover and a helicopter).
Finally, we plan to extend our approach to 3D mapping that capture richer scene representations beyond traversability maps.
}{
Future work includes expanding to 3D mapping for enhanced detail, personalizing learning for specific rover needs, and integrating inpainting and occupancy prediction to improve map quality and range.
}
\vspace{-0.25em}





\iftoggle{urs}{}

\iftoggle{doubleblind}{}{
\section*{ACKNOWLEDGMENT}
The authors thank Preston Culbertson, Jean-Pierre de la Croix, Georgios Georgakis, Shehryar Khattak, Jakob Eg Larsen, Mike Paton, and Niels H. Pontoppidan for their discussions during the development of this work.
}
\vspace{-0.5em}



\bibliography{ASL_bib,references}

@String { jrn_ACM_CACM              = {{Communications of the ACM}} }

@String { jrn_Elsevier_PCS          = {{Procedia Computer Science}} }

@String{jrn_IEEE_RAL              = {{IEEE Robotics and Automation Letters}}}

@String { jrn_IEEE_SPM              = {{IEEE Signal Processing Magazine}} }

@String{jrn_IEEE_TGRS                  = {{IEEE Transactions on Geosciences and Remote Sensing}}}

@String{jrn_IEEE_TIV                   = {{IEEE Transactions on Intelligent Vehicles}}}

@String { jrn_SAGE_IJRR             = {{Int.\ Journal of Robotics Research}} }

@String{jrn_Science_R                  = {{Science Robotics}}}

@String { proc_AAMAS                = {{Proc.\ Int.\ Conf.\ on Autonomous Agents and Multiagent Systems}} }

@String { proc_AISTATS              = {{AI \& Statistics}} }

@String { proc_ICLR                 = {{Int.\ Conf.\ on Learning Representations}} }

@String { proc_ICML                 = {{Int.\ Conf.\ on Machine Learning}} }

@String { proc_IEEE_AC              = {{IEEE Aerospace Conference}} }

@String { proc_IEEE_CVPR            = {{IEEE Conf.\ on Computer Vision and Pattern Recognition}} }

@String { proc_IEEE_ICRA            = {{Proc.\ IEEE Conf.\ on Robotics and Automation}} }

@String { proc_IEEE_IROS            = {{IEEE/RSJ Int.\ Conf.\ on Intelligent Robots \& Systems}} }

@String{proc_NeurIPS                = {{Conf.\ on Neural Information Processing Systems}}}

@String{proc_AIAA_Scitech              = {{AIAA Scitech Forum}}}

@Online{HoldenDayoubEtAl2023,
  author    = {Holden, L. and Dayoub, F. and Harvey, D. and Chin, T.-J.},
  title     = {Federated {Neural} {Radiance} {Fields}},
  year      = {2023},
  note      = {{Available at }\url{https://arxiv.org/pdf/2305.01163.pdf}},
}

@Online{Suzuki2023,
  author    = {Suzuki, T.},
  title     = {Federated Learning for Large-Scale Scene Modeling
with {Neural} {Radiance} {Fields}},
  year      = {2023},
  note      = {{Available at }\url{https://arxiv.org/pdf/2309.06030.pdf}},
}

@Article{LiSahuEtAl2020,
  author    = {Li, T. and Sahu, A. K. and Talwalkar, A. and Smith, V.},
  title     = {Federated Learning: Challenges, methods, and future directions},
  journal   = jrn_IEEE_SPM,
  volume    = {37},
  number    = {3},
  pages     = {50--60},
  year      = {2020},
}

@Inproceedings{McMahanMooreEtAl2017,
  author    = {{McMahan}, H. B. and Moore, E. and Ramage, D. and Hampson, S. and {Aguera} {y} {Arcas}, B.},
  title     = {Communication-Efficient Learning of Deep Networks
from Decentralized Data},
  booktitle = proc_AISTATS,
  year      = {2017},
}

@Article{MildenHallSrinivasanEtAl2021,
  author    = {Mildenhall, B. and Srinivasan, P. P. and Tancik, M. and Barron, J. T. and Ramamoorthi, R. and Ng, R.},
  title     = {{NeRF}: representing scenes as neural radiance fields for view synthesis},
  journal   = jrn_ACM_CACM,
  volume    = {65},
  number    = {1},
  pages     = {99--106},
  year      = {2021},
}

@Article{SchusterMuellerEtAl2020,
  author    = {Schuster, M. J. and M{\"u}ller, M. G. and Brunner, S. G. and Lehner, H. and {others}},
  title     = {The {ARCHES} Space-Analogue Demonstration Mission: Towards Heterogeneous Teams of Autonomous Robots for Collaborative Scientific Sampling in Planetary Exploration},
  journal   = jrn_IEEE_RAL,
  volume    = {5},
  number    = {4},
  pages     = {5315--5322},
  year      = {2020},
}

@Inproceedings{TancikSrinivasanEtAl2020,
  author       = {Tancik, M. and Srinivasan, P. P. and Mildenhall, B. and Fridovich-Keil, S. and Raghavan, N. and Singhal, U. and Ramamoorthi, R. and Barron, J. and Ng, R.},
  title        = {{Fourier} features let networks learn high frequency functions in low dimensional domains},
  booktitle    = proc_NeurIPS,
  year         = {2020},
}

@inproceedings{RahamanBaratinEtAl2019,
  title={On the Spectral Bias of Neural Networks},
  author={Rahaman, N. and Baratin, A. and Arpit, D. and Draxler, F. and Lin, M. and Hamprecht, F. and Bengio, Y. and Courville, A.},
  booktitle                = proc_ICML,
  year={2019},
}

@inproceedings{KwonParkEtAl2023,
  title={Renderable Neural Radiance Map for Visual Navigation},
  author={Kwon, O. and Park, J. and Oh, S.},
  booktitle=proc_IEEE_CVPR,
  year={2023}
}

@Online{NicholSchulman2018,
  author    = {Nichol, A. and Schulman, J.},
  title     = {Reptile: a scalable metalearning algorithm},
  year      = {2018},
  note      = {{Available at }\url{https://arxiv.org/pdf/1803.02999.pdf}},
}

@Inproceedings{ThakkerPatonEtAl2024,
  author        = {Thakker, R. and Paton, M. and Jones, B. and Daddi, G. and Royce, R. and Swan, M. R. and Swan, M. and {others}},
  title       = {To Boldly Go Where No Robots Have Gone Before - Part 4: {NEO} Autonomy for Robustly Exploring Unknown, Extreme Environments with Versatile Robots},
  booktitle    = proc_AIAA_Scitech,
  year         = {2024},
}

@Article{GeigerLenzEtAl2013,
  Title                    = {Vision meets robotics: The {KITTI} dataset},
  Author                   = {Geiger, A. and Lenz, P. and Stiller, C. and Urtasun, R.},
  Journal                  = jrn_SAGE_IJRR,
  Year                     = {2013},
  Volume                   = {32},
  Number                   = {11},
  Pages                    = {1231--1237},
}

@Inproceedings{TancikMildenHallEtAl2021,
  author    = {Tancik, M. and Mildenhall, B. and Wang, T. and Schmidt, D. and Srinivasan, P. P. and Barron, J. T. and Ng, R.},
  title     = {Learned initializations for optimizing coordinate-based neural representations},
  booktitle = proc_IEEE_CVPR,
  year      = {2021},
}

@Inproceedings{KingmaBa2015,
author    = {Kingma, D. P. and Ba, J. L.},
title     = {Adam: A method for stochastic optimization},
year      = {2015},
booktitle = proc_ICLR,
}

@Online{Agarap2018,
  author    = {Agarap, A. B.},
  title     = {Deep learning using rectified linear units ({ReLU})},
  year      = {2018},
  note      = {{Available at }\url{https://arxiv.org/pdf/1803.08375.pdf}},
}

@Online{ReddiCharlesEtAl2021,
  author    = {Reddi, S. and Charles, Z. and Zaheer, M. and Garrett, Z. and Rush, K. and Kone{\v{c}}n{\`y}, J. and Kumar, S. and {McMahan}, H. B.},
  title     = {Adaptive federated optimization},
  year      = {2021},
  note      = {{Available at }\url{https://arxiv.org/pdf/2003.00295.pdf}},
}

@Online{McMahanMooreEtAl2016,
  author    = {{McMahan}, H. B. and Moore, E. and Ramage, D. and {Ag{\"u}era} {y} {Arcas}, A.},
  title     = {Federated Learning of Deep Networks using Model Averaging},
  year      = {2016},
  note      = {{Available at }\url{https://arxiv.org/pdf/1602.05629v1/1000}},
}

@Inproceedings{DeLaCroixRossiEtAl2024,
  author       = {{de} {la} {Croix}, J.-P. and Rossi, F. and Brockers, R. and Aguilar, D. and Albee, K. and Boroson, E. and Cauligi, A. and {others}},
  title        = {Multi-Agent Autonomy for Space Exploration on the {CADRE} {Lunar} Technology Demonstration Mission},
  booktitle    = proc_IEEE_AC,
  year         = {2024},
}

@Online{ZhouPuEtAl2020,
  title={Distilled One-Shot Federated Learning},
  author={Zhou, Y. and Pu, G. and Ma, X. and Li, X. and Wu, D.},
  year      = {2021},
  note      = {{Available at }\url{https://arxiv.org/pdf/2009.07999.pdf}},
}

@Inproceedings{ParkHanEtAl2021,
  author       = {Park, Y. and Han, D.-J. and Kim, D.-Y. and Seo, J. and Moon, J.},
  title        = {Few-Round Learning for Federated Learning},
  booktitle    = proc_NeurIPS,
  year         = {2021},
}

@Inproceedings{SitzmannMartetEtAl2020,
  author       = {Sitzmann, V. and Martel, J. N. P. and Bergman, A. W. and Lindell, D. B. and Wetzstein, G.},
  title        = {Implicit neural representations with periodic activation functions},
  booktitle    = proc_NeurIPS,
  year         = {2020},
}

@Article{XiaHuEtAl2017,
author    = {Xia, G-.S. and Hu, J. and Hu, F. and Shi, B. and Bai, X. and Zhong, Y. and Zhang, L. and Lu, X.},
title     = {{AID}: A Benchmark Data Set for Performance Evaluation of Aerial Scene Classification},
year      = {2011},
journal   = jrn_IEEE_TGRS,
volume    = {55},
number    = {7},
pages     = {3965--3981},
}

@inproceedings{TianChangEtAl2023,
  title={Resilient and distributed multi-robot visual {SLAM}: Datasets, experiments, and lessons learned},
  author={Tian, Y. and Chang, Y. and Quang, L. and Schang, A. and {Nieto}-{Granda}, C. and How, J. P. and Carlone, L.},
  booktitle=proc_IEEE_IROS,
  year={2023},
}

@Online{GeyerKassahunEtAl2020,
  title={{A2d2}: {Audi} autonomous driving dataset},
  author={Geyer, J. and Kassahun, Y. and Mahmudi, M. and Ricou, X. and Durgesh, R. and Chung, A. S. and Hauswald, L. and Pham, V. H. and M{\"u}hlegg, M. and Dorn, S. and others},
  year      = {2020},
  note      = {{Available at }\url{https://arxiv.org/pdf/2004.06320.pdf}},
}

@Inproceedings{SunKretzschmarEtAl2020,
  author    = {Sun, P. and Kretzschmar, H. and Dotiwalla, X. and Chouard, A. and Patnaik, V. and Tsui, P. and Guo, J. and Zhou, Y. and Chai, Y. and Caine, B. and others},
  title     = {Scalability in perception for autonomous driving: {Waymo Open Dataset}},
  booktitle = proc_IEEE_CVPR,
  year      = {2020},
}

@Article{WilhelmGeisEtAl2020,
author    = {Wilhelm, T. and Geis, M. and P{\"u}ttschneider, J. and Sievernich, T. and Weber, T. and Wohlfarth, K. and W{\"o}hler, C.},
title     = {{DoMars16k}: A Diverse Dataset for Weakly Supervised Geomorphologic Analysis on {Mars} },
year      = {2020},
journal   = {Remote Sensing},
volume    = {12},
number    = {23},
pages     = {3981},
}

@Article{VermaMaimoneEtAl2023,
  Title                    = {Autonomous robotics is driving {Perseverance} rover’s progress on {Mars}},
  Author                   = {Verma, V. and Maimone, M. W. and Gaines, D. M. and Francis, R. and Estlin, T. A. and others},
  Journal                  = jrn_Science_R,
  Year                     = {2023},

  Number                   = {80},
  Pages                    = {1--12},
  Volume                   = {8},
}

@inproceedings{SaboiaRossiEtAl2024,
  title={{CADRE} {MoonDB}: Distributed Database for Multi-Robot
Information-Sharing and Map-Merging for Lunar Exploration},
  author={Saboia, M. and Rossi, F. and Nguyen, V. and Lim, G. and Aguilar, D. and {de} {la} {Croix}, J.-P.},
  booktitle=proc_AAMAS,
  year={2024},
}

@Inproceedings{WernerProencaEtAl2024,
  author       = {Werner, L. and Proen{\c c}a, P. and N{\"u}chter, A. and Brockers, R.},
  title        = {Covariance based terrain mapping for autonomous mobile robots},
  booktitle    = proc_IEEE_ICRA,
  year         = {2024},
}

@Inproceedings{AndersonBrownEtAl2024,
  author       = {Anderson, J. L. and Brown, T. L. and Cacan, M. and Kubiak, G. and Jasour, A. and Rothenberger, N. Z},
  title        = {Lessons from {Ingenuity's} Climb Up {Jezero} {Crater} {Delta}},
  booktitle    = proc_IEEE_AC,
  year         = {2024},
}

@Inproceedings{WangZhangEtAl2024,
author    = {Wang, L. and Zhang, K. and Zhou, A. and Simchowitz, M. and Tedrake, R.},
title     = {Robot Fleet Learning via Policy Merging},
year      = {2024},
booktitle = proc_ICLR,
}

@Online{NakanoyaImEtAl2021,
  title={Personalized Federated Learning of Driver Prediction Models for Autonomous Driving},
  author={Nakanoya, M. and Im, J. and Qiu, H. and Katti, S. and Pavone, M. and Chinchali, S.},
  year      = {2021},
  note      = {{Available at }\url{https://arxiv.org/pdf/2112.00956.pdf}},
}

@Article{XianjiaQueraltaEtAl2021,
  Title                    = {Federated Learning in Robotic and Autonomous Systems},
  Author                   = {Xianjia, Y. and Queralta, J. P. and Heikkonen, J. and Westerlund, T.},
  Journal                  = jrn_Elsevier_PCS,
  Year                     = {2021},
  Volume                   = {191},
  Pages                    = {135--142},
}

@article{qian2022hy,
  title={Hy-Seg: A hybrid method for ground segmentation using point clouds},
  author={Qian, Y. and Wang, X. and Chen, Z. and Wang, C. and Yang, M.},
  journal=jrn_IEEE_TIV,
  volume={8},
  number={2},
  pages={1597--1606},
  year={2022},
}

@article{zaheer2018adaptive,
  title={Adaptive methods for nonconvex optimization},
  author={Zaheer, M. and Reddi, S. and Sachan, D. and Kale, S. and Kumar, S.},
  journal=proc_NeurIPS,
  year={2018}
}

@inproceedings{proencca2022optimizing,
  title={Optimizing terrain mapping and landing site detection for autonomous {UAVs}},
  author={Proen{\c{c}}a, P. F. and Delaune, J. and Brockers, R.},
  booktitle=proc_IEEE_ICRA,
  year={2022},
}

@article{lin2017deep,
  title={Deep gradient compression: Reducing the communication bandwidth for distributed training},
  author={Lin, Yujun and Han, Song and Mao, Huizi and Wang, Yu and Dally, William J},
  journal={arXiv preprint arXiv:1712.01887},
  year={2017}
}
\bibliographystyle{ieeetr}

\end{document}